\documentclass[11pt,a4paper]{article}
\usepackage[hyperref]{eacl2021}
\usepackage{times}
\usepackage{latexsym}

\usepackage{microtype}

\aclfinalcopy 
\def\aclpaperid{121} 

\usepackage[T1]{fontenc}
\usepackage{type1cm}
\usepackage[american]{babel}
\usepackage{url}
\usepackage{amssymb} 
\usepackage{amsmath}
\usepackage{multirow} 
\usepackage{xspace}
\DeclareMathAlphabet{\mathcalbf}{OMS}{pzc}{b}{n}
\usepackage{microtype}

\usepackage[pdftex]{graphicx}
\usepackage{tabularx}
\usepackage{booktabs}
\usepackage{subcaption}
\DeclareGraphicsExtensions{.pdf,.ai,.jpg,.png}
\setkeys{Gin}{pagebox=artbox}

\graphicspath{{./}}
\newcommand{\bsfigure}[3][]{%
	\begin{figure}[t]
		\centering
		\includegraphics[#1]{#2}
		\caption{#3}\label{#2}%
	\end{figure}
}

\RequirePackage{type1cm}
\RequirePackage{color}
\RequirePackage{soul}
\setstcolor{blue}

\newsavebox\bscombox
\newcommand{\bscom}[3][]{%
  \sbox{\bscombox}{\fontsize{8}{9}\selectfont#1#2#3}
  \noindent
  \st{#2}{\selectfont
    \color{blue}#3\ifx\\#1\\\else{\fontsize{8}{9}\selectfont\color{violet}[#1]}\fi
    }
  }

\begin{document}

\title{Belief-based Generation of Argumentative Claims}

\author{
  	Milad Alshomary \quad Wei-Fan Chen \quad Timon Gurcke \quad Henning Wachsmuth \\
  	{\tt<first name>.<last name>@upb.de} \\
  	Department of Computer Science\hspace*{1em} \\
	Paderborn University, Paderborn, Germany
}

\date{}
\maketitle

\begin{abstract}
When engaging in an argumentative discourse, skilled human debaters tailor claims to the beliefs of the audience, to construct effective arguments. Recently, the field of computational argumentation witnessed extensive effort to address the automatic generation of arguments. However, existing approaches do not perform any audience-specific adaptation. In this work, we aim to bridge this gap by studying the task of {\em belief-based claim generation:} Given a controversial topic and a set of beliefs, generate an argumentative claim tailored to the beliefs. To tackle this task, we model the people's prior beliefs through their stances on controversial topics, and extend state-of-the-art text generation models to generate claims conditioned on the beliefs. Our automatic evaluation confirms the ability of our approach to adapt claims to a set of given beliefs. In a manual study, we additionally evaluate the generated claims in terms of informativeness and their likelihood to be uttered by someone with a respective belief. Our results reveal the limitations of modeling users' beliefs based on their stances, but demonstrate the potential of encoding beliefs into argumentative texts, laying the ground for future exploration of audience reach.
\end{abstract}

\section{Introduction}
\label{sec:introduction}

According to \newcite{van:1999}, debaters engaging in an argumentative discourse, aimed to resolve disagreement, design their next argumentative move considering the topical potential, the audience demand, and appropriate presentational devices. \newcite{feinberg:2015} stress based on the moral foundation theory \cite{godden:2010} how phrasing arguments to fit the audience's morals leads to a better agreement.
For example, in a debate on former US president \textit{Donald Trump}, potential topics could have been {\em immigration}, {\em health care plans}, {\em tax plans}, etc. However, knowledge about the audience being middle-class workers would have suggested to restrict the selection to \textit{Trump's tax plans}. An appropriate usage of presentational devices may have then put a con argument as follows: 

\paragraph{Example}
{\em ``Donald Trump was a bad president. He did nothing but hurt the poor and middle class, his tax plan benefited only rich people who could afford it.''}

\medskip
There is a recent growth of interest in argument generation as a subfield of computational argumentation. Several tasks have been proposed, including claim negation \cite{bilu:2015, hidey:2019}, counterargument generation \cite{hua:2019}, and conclusion generation \cite{alshomary:2020a}. While some research considers argumentative strategies when delivering arguments \cite{wachsmuth:2018b,el-baff:2019}, no one has worked on adapting arguments to user beliefs yet. Our goal is to bridge this gap. 

In this work, we propose to extend argument generation technologies with the ability to encode beliefs. This does not only better reflect the process by which humans reason, but it also allows controlling the output, in order to better reach the audience. In particular, we introduce the task of {\em belief-based claim generation:} Given a controversial topic and a representation of a user's beliefs, generate a claim that is both relevant to the topic and matches the beliefs. 

To approach this task, we first model user beliefs by their stances (pro or con) on a set of controversial topics, and then extend two state-of-the-art text generation approaches by conditioning their output on a specific set of beliefs. One approach builds on \newcite{li:2016}, equipping a sequence-to-sequence (Seq2Seq) model with a context vector representing the given stances. The other approach controls the output of a pre-trained argumentative language model (LM) using the algorithm of \newcite{dathathri:2020} to assure resembling the user's beliefs. We study the given task empirically on the \textit{debate.org} dataset of \newcite{durmus:2018}. The dataset contains users' arguments on various controversial topics as well as their stances towards the most popular topics on the website, called the \textit{big issues}. For our purposes, we use these big issues as the controversial topics, and we model beliefs by the user's stances towards them.

In our automatic evaluation, we compare both models against their unconditioned correspondents (i.e., the same models without knowledge about a user). We assess the generated claims in terms of the similarity to the ground truth and the likelihood of carrying textual features that reflect users' stances on big issues. Our results suggest that using users' beliefs significantly increases the effectiveness of the Seq2Seq and LM in most cases. Moreover, a stance classifier trained on claims generated by the conditioned LM achieves the best averaged accuracy across all big issues.

In a subsequent manual evaluation, we find that claims generated by the conditioned LM are more informative regarding the topic. In terms of predicting stance from generated claims, we analyze the limitations of our approach in detail, which lie in the belief encoding step. By avoiding these limitations, we find that the generated claims enable the annotators to predict correctly a stance on a given big issue in 45\% of the cases (26\% incorrectly). These results demonstrate the applicability of encoding a user's beliefs into argumentative texts, enabling future research on the effect of belief-based argumentative claims on audiences.

The contribution of this work is threefold\footnote{Code can be found under: \url{http://www.github.com/webis-de/eacl21-belief-based-claim-generation}}:
\begin{itemize}
	\setlength{\itemsep}{0pt}
	\item A new task, \textit{belief-based claim generation}.
	\item An approach to model and match users' beliefs in the generation of arguments.
	\item Empirical evidence of the applicability of encoding beliefs into argumentative texts.
\end{itemize}

\section{Related Work}
\label{sec:related-work}

Early research on argument generation aimed to create argumentative texts starting from a symbolic representation \cite{zukerman:2000,grasso:2000,carenini:2006}. Conceptually, those approaches all had a similar architecture consisting of three main phases: text planning, sentence planning, and realization \cite{stede:2018}. While they included a user model to a certain extent and aimed to generate convincing arguments, they were still performed on a limited scale.

With the tremendous advances of NLP and machine learning since then, research has begun to address different tasks in the realm of argument generation, showing promising results. \newcite{hua:2019} proposed a neural network-based framework for generating counter-arguments. Both \newcite{bilu:2015} and \newcite{hidey:2019} addressed the task of claim negation, using a rule-based and a neural approach respectively. Also, \newcite{sato:2015} proposed an approach to argument generation based on sentence retrieval, in which, given a topic, a set of paragraphs covering different aspects is generated. However, these approaches are agnostic to the target audience. 

\newcite{chen:2018a} modified the political bias of (often claim-like) news headlines using style transfer, accounting for general political sides (left and right) at least. Moreover, \newcite{wachsmuth:2018b} modeled rhetorical strategies in argument synthesis conceptually, but its computational realization \cite{el-baff:2019} considers the audience implicitly only, using a language model approach to select and arrange argumentative discourse units that are phrased in an argument. 

In the field of conversational AI, researchers have utilized machine translation techniques to tackle the task of dialog generation \cite{ritter:2011}. \newcite{li:2016} worked on augmenting sequence-to-sequence models by learning persona vectors from the given data. In a similar fashion, one of our approaches extends such a model by a context vector representing a user's belief. Here, however, we deal with argumentative text.

Progress in the field of text generation has been made due to the availability of large pre-trained language models \cite{devlin:2018, solaiman:2019}. While these models excel in generating coherent texts, ensuring a generated text possesses a certain property is not straightforward. Some research tackled this limitation, offering ways to better control the output \cite{keskar:2019,ziegler:2019}. One of the most flexible of such approaches is by \newcite{dathathri:2020}, which does not require fine-tuning for each controlling theme. Their algorithm conditions the output of a language model to contain certain properties defined by a discriminative classifier or a bag-of-words. One of our approaches makes use of this algorithm to condition the output of an argumentative language model on a bag-of-words that represents a user's beliefs. A recent relevant work by \newcite{schiller:2020} deals with the generation of aspect-controlled arguments. Similar to us, the authors utilize a pre-trained language model to generate arguments on a specific topic, with a controlled stance and aspect. Their focus is on topical aspects of arguments, though, and their approach based on \newcite{keskar:2019} is limited to a predefined set of topics and aspects. 
\section{Task} 
\label{sec:task-and-data}
Due to the importance of audience in argumentation when aiming for persuasiveness \cite{van:1999}, and due to the fact that humans comply to certain morals that shape their beliefs and affect their reasoning \cite{godden:2010, feinberg:2015}, we introduce the audience's beliefs as a new dimension to the argument generation process in this work. For this, we propose a new task, belief-based claim generation:

\begin{quote}
{\em Given a controversial topic and a representation of the audience's beliefs, generate a claim that is both relevant to the topic and matches the beliefs.}
\end{quote}

We focus this task on generating {\em claims} rather than full arguments to keep it simple and because claims denote the main units from which arguments are built. As shown by \newcite{feinberg:2015}, better agreement is achieved when arguments are framed with respect to audience's beliefs. Therefore, we argue that studying the mentioned task will enable argumentation technology, knowing its audience, to generate more convincing arguments, bridging the gap between disagreeing parties.

\subsection{Data} 

To study the proposed task, a dataset is needed in which information about users revealing their beliefs as well as their arguments on various topics are given. Here, we build upon the dataset introduced by \newcite{durmus:2018}, which was collected from \textit{debate.org}, an online platform where users can engage in debates over controversial topics and share their profiles. The dataset contains users' arguments as answers to topic questions and engagement in debates, along with various user information, including a user's self-specified stances (pro or con) on up to 48 predefined popular controversial topics, called {\em big issues}. 

\begin{table}[t!]%
\centering%
\small
\renewcommand{\arraystretch}{1}
\setlength{\tabcolsep}{7pt}%
\begin{tabular}{lrrr}
\toprule
\bf Dataset	& \bf \# Claims	& \bf \# Topics	& \bf \# Users	\\
\midrule
Training set 	& 41\,288 		& 22\,241  	& 5\,189		\\
Validation set 	& 5\,028 		& 2\,450 		& 2\,509 		\\
Test set 		&5\,154  		& 2\,728 		& 2\,512  		\\
\midrule
Full dataset	& 51\,470		&  27\,419		& 5\,189		\\
\bottomrule
\end{tabular}%
\caption{Number of claims, topics, and users in each of the training, validation, and test set of the data used in this paper.}
\label{table-data-stats}%
\end{table}

In our dataset, for the task at hand, we keep only users who have at least three arguments and stated their stance on at least one of the big issues. For those, we collected their arguments along with the topics and stances. In total, the dataset contains around 51k claims, on 27k topics from 5k users. We randomly split the dataset per topic into 10\% test and 90\% training. 10\% of the latter are used as the validation set. Statistics are given in Table~\ref{table-data-stats}.

To develop approaches to the belief-based claim generation task, we need training data where claims can be identified as such. Since claim detection is not our focus, we preprocess all data using the claim detection approach of \newcite{chakrabarty:2019}. In particular, we score the likelihood of each sentence being a claim, and only keep the one with the highest score as the user's claim on the topic. To evaluate the model, we created a sample of 100 arguments, and two annotators decided whether the extracted sentence represents a claim on the given topic or not. In terms of full agreement, the model extracted claims correctly in 81\% of the cases, the Cohen's $\kappa$ inter-annotator agreement being 0.3. We note that this preprocessing step produces some noise in the data, mainly affecting the training of our Seq2Seq model below.
\section{Approach}
\label{approach-section}

To study our research question, we propose and compare two approaches that build on top of known techniques for text generation. Both approaches rely on modeling users' beliefs via their stances on big issues. The first is an extension of the Seq2Seq model \cite{sutskever:2014}, where the user's stances are encoded as a context vector, while the second conditions the output of a pre-trained argumentative language model via a bag-of-words, constructed based on stances on big issues.

\subsection{Seq2Seq-based Model}

Given a topic, as a sequences of words $T = (w_1, w_2, ..., w_n)$, a user vector $ \overrightarrow{U} \in \{0, 1\}^k $ with $k$ being the number of big issues, and a claim as a sequence of words $ C = (w_{1}, w_{2}, ..., w_{m})$, first an LSTM-based encoder consumes the input topic and produces a hidden state $ \overrightarrow{h} $, which is used to initialize the LSTM-based decoder. The user vector $\overrightarrow{U}$ is projected into a new embedding space via a feed forward network with a learned weight matrix $W_U$, producing a new vector, $\overrightarrow{V}$:
$$
\overrightarrow{V} = \sigma ( W_U \cdot  \overrightarrow{U})
$$

%

Following \newcite{li:2016}, our $\overrightarrow{V}$ is served as their speaker embedding in the model. The difference between the speaker model in \citet{li:2016} and this model is that the vector $\overrightarrow{V}$ is not explicitly predefined but rather learned from the data, while in our model it is already predefined as a binary vector representing the user's stances on big issues. %
By augmenting the Seq2Seq model with a context user vector, the model is supposed to capture the correlation between users' stances on big issue and the corresponding claims. Once the correlation is learned, the model can generate a claim utilizing not only the topic, but also the stances on big issues of the target user, which reflect the beliefs.

\subsection{Conditioned Language Model}

In this approach, we represent a user's stances on big issues as a bag-of-words. We then use the topic as a prompt for a pre-trained argumentative language model (LM) to synthesize a claim conditioned using the algorithm of \newcite{dathathri:2020}. The synthesis process is illustrated in Figure~\ref{pplm-approach.pdf}.

\paragraph{Argumentative Language Model} 

Since we aim to generate {\em claims} in particular, a standard LM is not enough. To model argumentative language, we take a LM pre-trained on general language and fine-tune it on a large set of arguments (in our experiments, we use the corpus of \newcite{ajjour:2019}). The result is an LM that is able to generate argumentative text.

\bsfigure{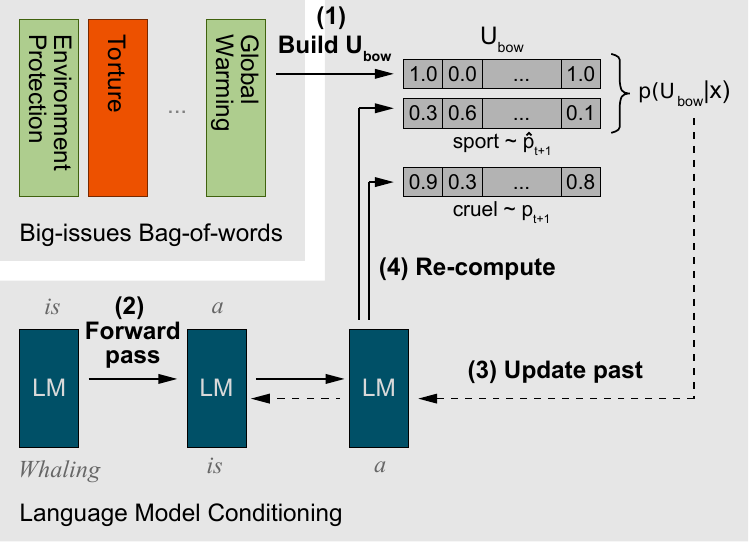}{The synthesis process of the conditioned LM on the topic ``Whaling'', given a user who is pro  {\em environmental protection} and {\em global warming} and con {\em torture}. Steps: (1) Building $U_{bow}$, based on stances (2) Forward pass through the LM to generate a token, {\em sport} (3) Updating the LM history $H_t$, based on $p(U_{bow}|x)$, and (4) Generating from the new history $ \hat{H}_t$ a new token {\em cruel}.}

\paragraph{Belief-based Bag-of-words} 

Next, we build a bag-of-words that represents the beliefs of a user. We learn this from the user's stances on the big issues. For example, a user pro {\em abortion} would likely be pro {\em choice}. Hence, words such as \textit{right} and \textit{choice} are candidates to be included in their belief-based bag-of-words. To this end, we first build two bag-of-words representations for each big issue, one for the pro and for the con side. For a user, we then construct a belief-based bag-of-words based on their stances on big issues.

To build a representative pro and con bag-of-words for each big issue, we follow the topic signature approach of \newcite{lin-hovy:2000}. Given a big issue, we first collect from some corpus of arguments three sets: relevant pro arguments $R_{pro}$, relevant con arguments $R_{con}$, and a random set of non-relevant arguments $\hat{R}$. For each relevant set, we then compute a likelihood ratio for all its words with respect to $\hat{R}$ and keep only words with a score higher than a specific threshold $\tau$, resulting in two sets of words, $W_{pro}$ and $W_{con}$. Since a word may appear in both sets, we remove it from the set where it occurs fewer times. Finally, we sort words according to their likelihood ratio and keep in both $W_{pro}$ and $W_{con}$ the top $k$ words, forming the final pro and con bag-of-words respectively.

\paragraph{Claim Generation} 

Given a user (represented by stances on big issues) and a topic, we construct a belief-based bag-of-words (Step 1 in Figure \ref{pplm-approach.pdf}): 
$$
U_{bow} = W_{1} \cup W_{2} \cup \ldots \cup W_{n} 
$$
where $W_{i}$ is the pro bag-of-words if the stance is pro and the con bag-of-words otherwise. Then, we use the topic as a prompt and the user's bag-of-words $U_{bow}$ to condition the generated claim (see Figure~\ref{pplm-approach.pdf}). In particular, given a transformer-based LM \cite{vaswani:2017}, a token $x_{t+1}$ is generated at each time step as follows:
$$
o_{t+1}, H_{t+1} = LM(x_t, H_t)
$$
$$
x_{t+1} \sim p_{t+1} = Softmax(W \cdot o_{t+1})
$$
where $H_t$ represents the history of the LM. Using the algorithm of \newcite{radford:2019}, called Plug and Play LM (PPLM), an update to the past, $\Delta H$, is computed to control the generated claim, based on the sum of the log likelihood $p(U_{bow}|x)$ of all words in the belief-based bag-of-words. Then the new history, $\hat{H}_t = H_t + \Delta H_t$, is used as in the previous equations to draw a new distribution $ \hat{p}_{t+1}$, of which a new token is sampled. To ensure fluency in the generated text, $\Delta H$ is further modified to ensure a high log-likelihood $p(x)$ with respect to the LM. More details on the algorithm can be found in the work of \newcite{radford:2019}.

In short, through fine-tuning an LM on argumentative text, we tune it to generate claims. Using the topic as a prompt, we ensure that the claim is on the topic. Finally, the PPLM represents beliefs, modeled as a bag-of-words $U_{bow}$, in the claim.

\section{Automatic Evaluation}

In this section, we evaluate whether utilizing user's beliefs as input, modeled as stances on big issues, leads to claims that better match the ground-truth claims and reveal the input stances on big issues.

\subsection{Experimental Setup}

On one hand, we compute the BLEU and METEOR scores of the generated claims with respect to the ground-truth claims. On the other hand, we compute the likelihood that the generated claims possess textual features that reflect the input user's beliefs. We do so by measuring the accuracy of predicting user's stances on big issues given the generated claims. We compute this accuracy for each of the 48 big issues individually and report the results for all of them. To this end, we carry out the following three steps for a given approach.

First, we generate claims for all given users and topics in the test dataset. Second, we keep only instances in which users have a stance (pro/con) on the tested big issue, and split the filtered dataset into training and test. Finally, we train a simple TF-IDF based linear classifier on the training set to predict the stance on the big issue given the text of the claim. The accuracy of the classifier on the test split then quantifies the likelihood of the generated claims possessing textual features that reflect the stance on the corresponding big issue.

\begin{table}[t!]%
\centering%
\small
\setlength{\tabcolsep}{5pt}%
\begin{tabular}{lrrr}
\toprule
\bf Approach & \bf BLEU-1 & \bf BLEU-3 & \bf METEOR\\
\midrule
S2S-baseline  & 18.2\%     &     0.44\% &  \bf 16\%\\
S2S-model     & \bf \textsuperscript{*}18.4\%  & \bf \textsuperscript{*}0.46\%  &  \bf 16\%\\
\addlinespace
LM-baseline   & \phantom{0}9.6\%      &  \bf 0.26\%  & \phantom{0}8\% \\
LM-conditioned    & \bf \textsuperscript{*}12.0\%  &  0.16\%  & \bf \textsuperscript{*}11\%\\
\bottomrule
\end{tabular}%
\caption{BLEU and METEOR scores of the claims of each evaluated approach compared to the ground-truth claims. Values marked with * are significantly better than the respective baseline at $p$ < .05 (student's $t$-test).}
\label{table-automatic-evaluation-1}%
\end{table}


\begin{table}[t!]%
\end{table}


\begin{table*}[t!]%
\centering%
\small
\renewcommand{\arraystretch}{1}
\setlength{\tabcolsep}{2.25pt}%
\begin{tabular}{l@{\hspace{-0.7em}}rrrrrrrrrr@{\hspace{1.0em}}r}
\toprule
\bf                 & \bf               & \bf Death & \bf Gay & \bf Drug & \bf Global & \bf Environm. & \bf Medical & \bf Smok. & \bf Minim. & \bf Border  & \bf All 48 \\
\bf Approach & \bf Abortion & \bf penalty & \bf Marriage & \bf legaliz. & \bf warming & \bf protection & \bf mariju. & \bf ban & \bf wage & \bf fence  & \bf big issues  \\
\midrule
Ground-truth          & 0.49 & 0.59 & 0.55 & 0.55 & 0.55 & 0.55 & 0.50 & 0.53 & 0.48 & 0.62 & 0.52 \\
\addlinespace
S2S-baseline& 0.49 & 0.48 & \bf 0.52 & 0.45 & 0.51 & 0.51 & 0.57 & \bf 0.53 & 0.53 & 0.46 & 0.50 \\
S2S-model   & 0.55 & \bf 0.55 & 0.45 & 0.45 & 0.51 & \bf 0.58 & 0.57 & \bf 0.53 & 0.49 & \bf 0.52 & 0.51 \\
\addlinespace
LM-baseline & 0.48 & 0.50  & 0.54 & 0.49 & 0.54 & 0.56 & 0.51 & 0.45 & 0.59 & 0.46 & 0.50 \\
LM-conditioned    & \bf \textsuperscript{*}0.58 & \textsuperscript{*}0.53 & 0.45 & \bf 0.56 & \bf \textsuperscript{*}0.61 & \bf 0.58 & \bf 0.58 & \bf 0.53 & \bf 0.65 & 0.50 & \bf 0.54\\
\midrule
\# Training & 1\,610 & 1\,532 & 2\,098 & 1\,538 & 1\,960 & 2\,196 & 2\,096 & 1\,370 & 1\,580 & 1\,092 & - \\
\# Test & 350 & 366 & 196 & 316 & 156 & 86 & 138 & 294 & 172 & 280 & - \\
\bottomrule
\end{tabular}%
\caption{Accuracy of each classifier trained on claims generated by the evaluated approaches to predict the stance, on the 10 most frequent big issues as well as on average over all 48 big issues. Values marked with * are significantly better than corresponding baseline at $p < 0.05$ according to a one-tailed Student's $t$-test.}
\label{table-automatic-evaluation-2}%
\end{table*}

\subsection{Implementation Details}

In the following, we give implementation details of our approaches and the corresponding baselines:

\paragraph{Seq2seq-based Model} 

Based on the OpenNMT framework \cite{klein:2017}, the encoder and decoder are each two-layer LSTMs of hidden size 512 with GloVe word embeddings of size 300. Users' stances on big issues are represented as a one-hot encoded vector, and then projected into 16 dimensions space through a one-layer dense neural network. We trained the model with the Adagrad optimizer (batch size 16) and refer to it as \textit{S2S-model}.

\paragraph{Conditioned Language Model} 

We constructed the pro/con relevant argument sets ($R_{pro} , R_{con}$) by querying the respective big issue from the API provided by \newcite{ajjour:2019} and extracting  pro/con arguments from the top 60 results. For the non-relevant argument set ($\hat{R}$), we used the same corpus \cite{ajjour:2019} and randomly selected 100 arguments. We eliminated all words with a score under $\tau = 10$ and finally kept the top $k = 25$ words from each set ($R_{pro} , R_{con}$) to represent the bag-of-words.\footnote{We refrained from tuning the parameters here since we do not have a ground truth.}

To model the argumentative language, we fine-tuned the GPT-2 model on the corpus of \newcite{ajjour:2019}, which contains around 400k arguments. The fine-tuning was performed using the transformers framework \cite{wolf:2019}. We used the topic as a prompt to trigger the generation process. However, since some topics are phrased as a question (e.g, ``is abortion wrong?''), we extracted the noun phrase from the topic and used it as a prompt. For conditioning the generated claim, we used the PPLM implementation \cite{dathathri:2020}\footnote{step-size=0.15 and the repetition-penalty=1.2}. We call this model the \textit{LM-conditioned}.

\paragraph{Baselines} 

To evaluate the gain of encoding user's beliefs, we compare our two approaches to the corresponding version without stances on big issues as an input. We refer to these baselines as \textit{S2S-baseline} and \textit{LM-baseline} respectively\footnote{A baseline that uses the corresponding bag-of-words of the targeted topic to guide the generation wouldn't be valid, since we don't have information on the user's stance on this targeted topic.}.

\subsection{Results} 

Table~\ref{table-automatic-evaluation-1} shows the results of our approaches and the baselines in terms of BLEU and METEOR. 

For \textit{S2S}, the BLEU scores of our approach are significantly better than the baseline. The \textit{LM-conditioned} is significantly better than the baseline version in terms of BLEU-1 and METEOR. In general, the S2S-model has the highest scores across all measures. The reason may be that it was trained in a supervised manner on the given dataset, whereas the \textit{LM-model} was only fine-tuned in an unsupervised way on a different argument corpus.

Regarding the encoding of user stances, Table~\ref{table-automatic-evaluation-2} shows the accuracy of a linear classifier trained to predict the stance from the claims generated by each approach as well as from the ground-truth, on average and on the 10 most frequent big issues. A complete table with all big issues can be found in the appendix. 

The best average accuracy across all the big issues is achieved by the \textit{LM-model} (0.54). Compared to the corresponding baselines, the LM-model and the \textit{S2S-model} generated claims that boosted the accuracy of the stance classifier on 33 (69\%) and 21 (44\%) of all big issues respectively. Overall, in 20 of the big issues, the best accuracy was achieved on the claims generated by the conditioned LM, compared to only nine big issues for the S2S-model. This indicates that the LM-conditioned can better encode a user's beliefs, modeled as stances on big issues, into generated claims.

\section{Manual Evaluation}

To obtain more insights into belief-based claim generation, we let users manually evaluate the output of the given approaches.
Upon inspecting a sample of generated claims by our approaches, we noticed that the \textit{LM-conditioned} produces more fluent and informative texts. Accordingly, we focused on the LM-conditioned and its baseline in the evaluation, where we conducted two user studies. The goal of the first was to assess the quality of the big-issue bag-of-words collected automatically, while the second targeted the output of the LM-model, its baseline, and a variant that utilizes a manually refined bag-of-words.

\subsection{Automatic Collection of Bag-of-words}

To keep the manual annotation effort manageable, we evaluated only the top-10 big issues. Two authors of this paper categorized each word in the pro/con bag-of-words of the corresponding big issue into five categories, {\em c1--c5}:
\begin{enumerate}
\setlength{\itemsep}{0pt}
\item[c1:] 
Word irrelevant to the big issue.
\item[c2:] 
Relevant word, wrong stance.
\item[c3:]
Relevant word, both stances possible.
\item[c4:]
Relevant word, correct stance.
\item[c5:]
Very relevant word, correct stance.
\end{enumerate}

\begin{table*}[t!]%
\small
\centering%
\renewcommand{\arraystretch}{1}
\setlength{\tabcolsep}{3.25pt}%
\begin{tabular}{lrrr@{\quad\quad}rrr@{\quad\quad}rrr@{\quad\quad}rrr}
\toprule
 & \multicolumn{3}{l}{\bf \quad\quad\, Overall} & \multicolumn{3}{l}{\bf Relatedness Level 4} & \multicolumn{3}{l}{\bf Relatedness Level 3} & \multicolumn{3}{c}{\bf Relatedness Level-2}\\
\cmidrule(l@{0pt}r@{16pt}){2-4}\cmidrule(l@{0pt}r@{16pt}){5-7}\cmidrule(l@{0pt}r@{16pt}){8-10}\cmidrule(l@{0pt}r@{2pt}){11-13}
\bf Approach						   & \bf True & \bf False & \bf Undec. & \bf True & \bf False & \bf Undec. & \bf True & \bf False & \bf Undec. & \bf True & \bf False & \bf Undec.\\
\midrule
LM-baseline	   & 44\% & 34\% & \bf 22\% &  50\% & 50\% & \bf 0\% &   55\% & 31\% & 14\% &  \bf 27\% & 20\% & \bf 53\% \\
LM-conditioned		   & 37\% & 32\% & 31\% &  35\% & 38\% & 27\% &  59\% & 41\% & \bf 0\% &   13\% & \bf 13\% & 74\% \\
LM-cond.\ (manual) & \bf 45\% & \bf 26\% & 28\% & \bf 50\% & \bf 31\% & 19\% &  \bf 61\% & \bf 28\% & 11\% &  25\% & 18\% & 56\% \\
\midrule
Ground Truth			   & 42\% & 30\% & 28\% &  38\% & 42\% & 19\% &  64\% & 27\% & 9\% &   27\% & 19\% & 54\% \\
\bottomrule
\end{tabular}%
\caption{Manual Evaluation: Percentage of cases for each approach where the majority of annotators predicted the stance of a generated claim on the given big issue correctly (true), incorrectly (false), or could not decide it (Undec.). The overall scores and those for each topic/big-issue relation level are listed.}%
\label{table-manual-evaluation}
\end{table*}

\begin{table}[t!]%
\small
\centering%
\renewcommand{\arraystretch}{1}
\setlength{\tabcolsep}{4pt}%
\begin{tabular}{lccccc}
\toprule
& \bf Irrelevant & \multicolumn{3}{c}{\bf Relevant} & \multicolumn{1}{c}{\bf Very Relevant} \\
\cmidrule(l@{0pt}r@{0pt}){2-2}\cmidrule(l@{5pt}r@{0pt}){3-5} \cmidrule(l@{5pt}r@{0pt}){6-6}
\bf Words & \bf c1 & \bf c2 & \bf c3  & \bf c4 & \bf c5\\
\midrule
Pro & 14\% & 10\% & 36\% & 34\% & 6\% \\
Con & 36\%& 2\% & 34\% & 26\% & 2\% \\
\bottomrule
\end{tabular}%
\caption{Distribution of the pro/con bag-of-words, averaged across the top-10 big issues, over the five considered categories: c2 means wrong stance, c3 words that fit both stances, and c3 and c4 represent correct stance.}
\label{table-bow-collection-evaluation-1}%
\end{table}


Examples can be found in the appendix. To compute inter-annotator agreement, three big issues were annotated by both annotators, resulting in Cohen's $\kappa$ of 0.45, reflecting moderate agreement. Afterwards, only one annotator continued the annotations for the other big issues. 

Table \ref{table-bow-collection-evaluation-1} shows the distribution of words over categories, averaged across the 10 big issues. For the pro bag-of-words, around 40\% of the words are relevant and reflect the right stance, while 36\% are relevant but could be used in arguments from both stances. For the con bag-of-words, however, the percentages are lower (28\% and 34\% respectively). A considerable proportion of words belong to categories c1 and c2, which creates noise that could confuse the conditioning process of the LM. Hence, we also consider a variant of the conditioned LM that uses only relevant words from c4 and c5.


\subsection{Claim Generation} 

We evaluate the effectiveness in terms of whether a given generated claim reveals the stance of the given user on a specific big issue as well as how informative the claim is regarding the given topic.

Since not all topics are directly related to the big issues that can be revealed in the generated claims, we manually annotated the relatedness of the top frequent 200 topics in the test dataset to the most frequent 10 big issues, and created the evaluation sample accordingly. In particular, two authors of this paper scored the relatedness of each pair of topics and big issues on a scale from 1 to 4: 
\begin{enumerate}
\setlength{\itemsep}{0pt}
\item[4:] 
Topic and big issue are the same. Example: {\em "gay marriage should be legalized"
 and "gay marriage"}
\item[3:] 
A stance on the topic likely affects the stance on the big issue. Example: {\em "killing domestic abusers" and "death penalty" 
}
\item[2:]
A stance on the topic may affect the stance on the big issue. Example: {\em "morality" and "abortion"}
\item[1:]
Topic and big issue are not related. Example: {\em "do aliens exist?" and "abortion"}
\end{enumerate}

The two annotators had a Cohen's $\kappa$ agreement of 0.54. Around 97.4\% of all pairs got score~1, 1.1\% score 2, 0.8\% score 3, and 0.7\% score 4. The small percentage of cases that can be evaluated reflects a limitation in the designed evaluation study. However, it still allows us to evaluate the effectiveness of our approach for different levels of relatedness. Given the annotated pairs, we randomly selected 10 pairs from levels 2, 3, and 4 each. For each pair, we then collected all claims on the topic from the test set, where the author specifies a stance on the corresponding big issue. We randomly select 30 claims each, resulting in an evaluation sample of 90 instances.

We used the crowdsourcing platform \textit{MTurk}\footnote{A crowd sourcing platform: \url{https://www.mturk.com/}} for evaluation. For each instance, we showed a topic, a claim, and the corresponding big issue to three annotators. The annotators had to perform two tasks:
(1)~to predict the stance of the user on the corresponding big issue from the text of the claim, and
(2)~to rate the claim's informativeness regarding the topic on a scale from 1 to 3. 

Table~\ref{table-manual-evaluation} shows the percentage of cases in which the majority of annotators predicted the stance correctly (true), incorrectly (false), or could not decide about the stance (undec.) from the generated claim. Across the whole sample (Overall), the claims generated by \textit{LM-conditioned (manual)}, the model conditioned on the refined bag-of-words, most often allowed to predict the stance correctly (45\%). We thus attribute the low effectiveness of the \textit{LM-model} to the noise generated by the automatic collection of big-issues' bag-of-words, especially seeing that the effectiveness gets better across all levels when eliminating this noise. 

Analyzing each relatedness level individually yields more insights. For relatedness level~4, where the topic is the same as the big issue, the \textit{LM-conditioned (manual)} generated claims where the majority of the cases with known stance were correct (63\%). In level~3, we observe that both versions of our approach outperform the baseline in producing claims that express the correct stance on the corresponding big issue with percentages of 59\% and 68\% respectively. Finally, at relation level 2, which represents a weak relation between topics and big issues, predicting the stance seems to become hard, as indicated by high percentages of undecided cases. We believe that the weak relatedness made the annotators guess the stance in some cases, leading to unreliable annotations.%

\begin{table}[t!]%
\small
\centering%
\renewcommand{\arraystretch}{1}
\setlength{\tabcolsep}{5pt}%
\begin{tabular}{l@{\!\!\!\!\!}rrrr}
\toprule
\bf Approach & \bf Overall & \bf Level 4 & \bf Level 3 & \bf Level 2\\
\midrule
LM-baseline	   & 1.8     & \bf 2.5& 1.9     & 1.4 \\
LM-conditioned & \bf 2.1 &  2.3   & \bf 2.5 & \bf 1.5  \\
LM-cond. (manual) & 2.0     &  2.3   & 2.2     & 1.5 \\
\midrule
Ground Truth			   & 2.0     &  1.9    & 1.8     & 2.2\\
\bottomrule
\end{tabular}%
\caption{Manual Evaluation: Mean informativeness of the claims generated by each approach with regard to the topic (1--3, higher is better). The overall scores and those for each topic/big-issue relation level are listed.}
\label{table-manual-evaluation-2}%
\end{table}

Table~\ref{table-manual-evaluation-2} shows the average score of all approaches regarding the informativeness of the generated claims. Here, both versions of our approach achieved better scores than the baseline, matching the ground-truth score. We believe that the low scores of the ground-truth claims stem from the noise generated in the claim detection step.


\begin{table}[t!]%
	\centering%
	\small
	\renewcommand{\arraystretch}{1.0}
	\setlength{\tabcolsep}{2.5pt}%
	\begin{tabular}{p{1.3cm}p{4.85cm}p{1.05cm}}
	
		\multicolumn{2}{l}{(a) {\em Topic:} is abortion ok} \\
		\multicolumn{2}{l}{\quad\, {\em Big issue:} (Con) Abortion} & {\em Level:} 4 \\
		\toprule
		\textbf{Approach} & \textbf{Claim} & \textbf{Stance} \\
		\midrule
		LM-cond. & abortion rights groups argue that the right to abortion is a fundamental human right. & Undec.\\
		\addlinespace
		LM-baseline & abortionists are not the only ones who are against abortion. There are many other people who are against abortion & Undec. \\
		\bottomrule
		
		\addlinespace
		\addlinespace
		\multicolumn{2}{l}{(b) {\em Topic:} abortion is right or wrong} \\
		\multicolumn{2}{l}{\quad\, {\em Big issue:} (Con) Abortion} & {\em Level:} 4 \\
		\toprule
		\textbf{Approach} & \textbf{Claim} & \textbf{Stance} \\
		\midrule
		LM-cond. & abortion right or wrong is a personal choice. I believe that abortion should be legal because it's the woman"s body and she has every right to do what ever she wants. & Pro\\
		\addlinespace
		LM-cond. (manual) & abortion right or wrong. I believe that abortion is murder and should be illegal in all cases except when the mother's life is at stake, rape, incest etc. & Con \\
		\bottomrule
		
		\addlinespace
		\addlinespace
		\multicolumn{2}{l}{(c) {\em Topic:} whaling} \\
		\multicolumn{2}{l}{\quad\, {\em Big issue:} (Pro) Environmental Protection} & {\em Level:} 3 \\
		\toprule
		\textbf{Approach} & \textbf{Claim} & \textbf{Stance} \\
		\midrule
		LM-cond. & whaling is a very dangerous and cruel practice. It causes great harm to the environment, animals that are caught in it have their lives cut short by being eaten alive. & Pro\\
		\addlinespace
		LM-baseline & whaling is a sport. It is a sport because it involves a lot of physical exertion and skill. & Undec. \\
		\bottomrule
		
		\addlinespace
		\addlinespace
		\multicolumn{2}{l}{(d) {\em Topic:} should murder be legalized?} \\
		\multicolumn{2}{l}{\quad\, {\em Big issue:} (Pro) Death Penalty} & {\em Level:} 2 \\
		\toprule
		\textbf{Approach} & \textbf{Claim} & \textbf{Stance} \\
		\midrule
		LM-cond. & should murder be legalized? I think that abortion should not be legal. Abortion is killing a baby, and it's wrong to kill an innocent human being. & Undec.\\
		\addlinespace
		LM-baseline & should murder be legalized? I think so. I think that it should be legalized because it is a good thing & Undec. \\
		\bottomrule
	\end{tabular}%
	\caption{A selection of claims generated by the different evaluated approaches for the different association levels between topic and big issue discussed in the text.}
	\label{table-example-claims}%
\end{table}

\paragraph{Error Analysis} 

Table~\ref{table-example-claims} shows some cases from our evaluation. In case (a), even though the model generated a claim focusing on \textit{fundamental rights}, the majority of the annotators was undecided about the stance. Case (b) shows the effect of noise in the bag-of-words: the generated claim reflects a pro stance on abortion, while the ground-truth stance is con. This is avoided in the claim generated by {\em LM-conditioned (manual)}. Case (c) shows a working example of which our approach correctly generated a claim on {\em whaling} from an environmental perspective when conditioned as such. Case (d) is a level~2 example, indicating limitation in our evaluation, namely, the generated claim reveals a stance on abortion, but we asked about death penalty.

\section{Conclusion}

In this paper, we have proposed to equip argument generation technology with the ability to encode beliefs for two reasons: first, it reflects the human process of synthesizing arguments, and second, it gives more control on the generated arguments leading to a better reach of the audience. For this purpose, we have presented the task of belief-based claim generation. Concretely, we studied the research questions of how to model a user's beliefs as well as how to encode them when generating an argumentative text. We have modeled users' beliefs via their stances on big issues, and used them as an extra input in our approaches.

Our automatic evaluation has provided evidence of the applicability of encoding beliefs into argumentative texts. In manual studies, we found that limitations in the effectiveness of our approach stem from noise produced by the automatic collection of a bag-of-words. The findings of this paper lay the ground to investigate the role of beliefs in generating arguments that reach their audience.

We point out that ethical issues arise, when tuning arguments to affect specific people, such as attempts to manipulate them. While the task and settings considered here are rather too fundamental to already make these issues critical, future work should pay attention to them. Our goal is to develop systems that bring people together.

\section*{Acknowledgments}

We thank the anonymous reviewers for their helpful feedback.
This work was partially supported by the German Research Foundation (DFG) within the Collaborative Research Center ``On-The-Fly Computing'' (SFB~901/3) under the project number~160364472.

\bibliography{eacl21-belief-based-claim-generation-lit}
\bibliographystyle{acl_natbib}

%
\end{document}